\UseRawInputEncoding
\documentclass[conference]{IEEEtran}
\IEEEoverridecommandlockouts
\usepackage{cite}
\usepackage{amsmath,amssymb,amsfonts}
\usepackage{algorithmic}
\usepackage{graphicx}
\usepackage{textcomp}
\usepackage{xcolor}
\usepackage{comment}
\usepackage{listings}
\usepackage{xcolor}
\usepackage{hyperref}
\usepackage{fontawesome}
\def\BibTeX{{\rm B\kern-.05em{\sc i\kern-.025em b}\kern-.08em
    T\kern-.1667em\lower.7ex\hbox{E}\kern-.125emX}}
\begin{document}

\title{NL2AGBench: Benchmarking LLM Auto-Formalization for AlphaGeometry\\
}

\author{
\IEEEauthorblockN{Samuel Xiao}
\IEEEauthorblockA{\textit{Valley Christian High School} \\
Fremont, CA \\
}
\and
\IEEEauthorblockN{Judy Song}
\IEEEauthorblockA{\textit{Vandegrift High School}\\
Austin, TX \\
}
\and
\IEEEauthorblockN{Rory Hu}
\IEEEauthorblockA{\textit{Groton School}\\
Cupertino, CA \\
}
\and
\IEEEauthorblockN{Ziliang Zong}
\IEEEauthorblockA{\textit{Computer Science Department} \\
\textit{Texas State University}\\
}
}

\maketitle

\begin{abstract}

Recent advances in large language models (LLMs) have demonstrated strong performance in natural language understanding and mathematical reasoning. However, their ability to translate informal mathematical problem statements into formal representations remains underexplored. This limitation is particularly important for neuro-symbolic geometry systems such as AlphaGeometry, whose theorem-proving engine requires inputs written in a specialized domain-specific language (DSL). Although AlphaGeometry achieved near-IMO gold-medalist performance on Olympiad geometry problems, the manual conversion of natural-language problems into its formal syntax remains a significant usability bottleneck. To address this gap, we introduce Natural Language to AlphaGeometry Benchmark (NL2AGBench), which aims to evaluate the ability of LLMs to translate English geometry problems into AlphaGeometry-compatible formal representations. NL2AGBench evaluates translation quality using execution-based verification within the AlphaGeometry framework rather than relying solely on textual similarity metrics. We evaluate ten state-of-the-art open- and closed-source LLMs spanning multiple parameter scales and analyze their performance in terms of executable translation accuracy, syntactic correctness, and error characteristics. Our experiments reveal a substantial performance gap between frontier closed-source models and open-source alternatives. While leading closed-source models achieve executable translation rates exceeding 80\%, even the largest open-source models struggle to consistently preserve geometric constraints and produce valid formalizations. To better understand these failures, we introduce an error taxonomy that categorizes AlphaGeometry execution failures into syntax and logic errors. We further investigate mitigation strategies, including few-shot prompting, fine-tuning, and human-guided hinting, demonstrating measurable improvements across multiple model families.

\end{abstract}

\begin{IEEEkeywords}
AlphaGeometry, benchmark, translation, neuro-symbolic learning, prompt engineering.
\end{IEEEkeywords}

\begin{figure*}[t] 
    \centering
    \includegraphics[width=.9\linewidth]{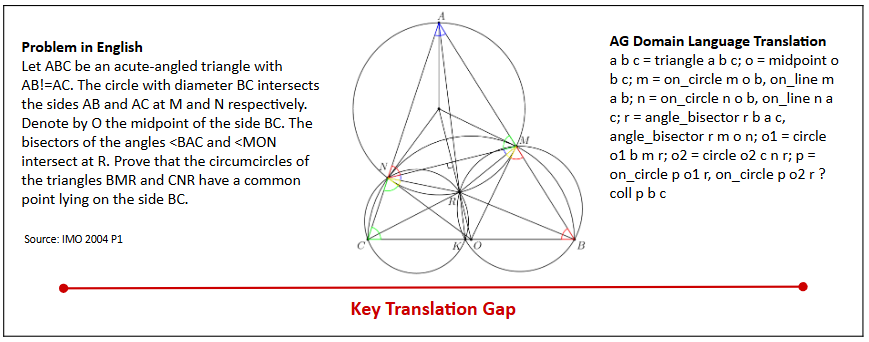}
    \caption{Translation gap between natural-language geometry problems and AlphaGeometry's formal DSL. While AlphaGeometry has demonstrated strong theorem-proving capabilities, the manual formalization step remains a major barrier to usability and large-scale deployment.}
    \label{fig:translation_difficulty}
\end{figure*}

\section{Introduction}

Recent progress in large language models (LLMs) has significantly advanced natural language understanding, code generation, and mathematical reasoning. Despite these advances, translating informal mathematical descriptions into machine-verifiable formal representations remains a challenging problem. This process, commonly referred to as auto-formalization, is a critical prerequisite for many theorem-proving and neuro-symbolic reasoning systems because formal solvers require precise symbolic inputs rather than natural-language descriptions.

Geometry theorem proving presents a particularly challenging auto-formalization setting. Unlike algebraic problems, geometry problems often contain implicit spatial relationships, diagram-dependent information, and complex geometric constraints that must be preserved exactly during translation. Even small translation errors can fundamentally alter a problem's meaning and prevent downstream reasoning systems from producing valid proofs.

AlphaGeometry represents one of the most successful neuro-symbolic systems for Olympiad-level geometry reasoning \cite{trinh2024alphageometry}. By combining neural-guided auxiliary construction generation with symbolic deductive reasoning, AlphaGeometry solved 25 of 30 historical International Mathematical Olympiad (IMO) geometry problems and demonstrated performance comparable to elite human competitors. However, AlphaGeometry requires all problems to be expressed in a specialized domain-specific language (DSL). As a result, geometry problems written in standard mathematical English must first be manually translated into formal syntax before theorem proving can begin. This translation requirement remains a major barrier to accessibility, scalability, and practical deployment. Figure 1 illustrates this challenge. On the left is a geometry problem written in natural mathematical language, similar to the format used in Olympiad competitions and textbooks. On the right is the corresponding AlphaGeometry DSL representation required by the theorem prover. While the English description is concise and intuitive for human readers, the formal representation must explicitly encode geometric objects, constraints, incidences, and proof targets using a strict symbolic syntax. Minor omissions or incorrect constructions can fundamentally change the problem semantics and prevent the theorem prover from generating a valid proof.

The challenge of converting natural-language geometry problems into formal representations is closely related to semantic parsing and mathematical auto-formalization. Prior work has explored translating informal mathematics into proof-assistant languages such as Lean, Isabelle, and Metamath, while recent studies have investigated the use of LLMs to assist formal theorem proving \cite{berant2013semantic,dong2016language}. However, existing benchmarks primarily evaluate proof generation, mathematical reasoning, or general auto-formalization. To the best of our knowledge, no benchmark specifically evaluates the translation of natural-language geometry problems into AlphaGeometry-compatible formal representations.

To address this gap, we introduce NL2AGBench, a benchmark designed to evaluate the ability of LLMs to translate English geometry problems into AlphaGeometry's DSL. NL2AGBench consists of geometry problems paired with verified formal representations and evaluates model outputs through direct execution within AlphaGeometry. This execution-based evaluation allows us to assess not only syntactic correctness but also whether generated translations preserve the geometric semantics required for theorem proving.

Using NL2AGBench, we evaluate a diverse collection of open- and closed-source LLMs and analyze their ability to bridge the gap between natural-language problem descriptions and formal geometric reasoning systems. Our results reveal a substantial performance disparity between frontier closed-source models and open-source alternatives, highlighting the difficulty of geometry-specific auto-formalization. We further characterize common failure modes through a systematic error taxonomy and analyze mitigation strategies including few-shot prompting, fine-tuning, and human-guided hints.

Our main contributions are summarized as follows:

\begin{itemize}
    \item We introduce NL2AGBench, a benchmark for evaluating the translation of natural-language geometry problems into AlphaGeometry-compatible domain-specific language representations. The benchmark provides a standardized framework for studying geometry auto-formalization and LLM-assisted theorem-proving pipelines.
    \item We conduct a comprehensive evaluation of ten state-of-the-art LLMs, including both open-source and closed-source models spanning multiple parameter scales. Our results reveal a substantial performance gap between frontier proprietary models and currently available open-source alternatives. The largest open-source model achieves a score of under 46\% with an optimized prompting technique, while closed-source large frontier models have reached 85\% and above.
    \item We introduce a systematic error taxonomy for AlphaGeometry translation failures, categorizing execution failures into syntax and logic errors. This taxonomy provides practical insights into common failure modes and facilitates future research on geometry auto-formalization.
    \item We investigate multiple translation-improvement strategies, including few-shot prompting, fine-tuning, and human-guided hinting. Our experiments demonstrate that targeted prompting and guidance can significantly improve pass rates for Llama3.1:70b by 23\%, GPT-4o-mini:8b by 8\%, and Qwen3:235b by 33.3\%.
\end{itemize}


\section{Related Work}

\subsection{AlphaGeometry and its Applications}

AlphaGeometry is a neuro-symbolic AI system designed to solve Euclidean geometry proof problems \cite{trinh2024alphageometry}. It combines a symbolic deduction engine with a neural model that proposes auxiliary constructions when direct deduction is insufficient. This builds on earlier symbolic geometry theorem-proving methods, such as automated geometry proof systems \cite{chou2000deductive}. Beyond pure mathematics, AlphaGeometry has potential applications in scientific reasoning, synthetic-data generation, and operating systems verification. Other recent systems, such as HAGeo and TongGeometry, have also explored improved auxiliary construction strategies and guided search for Olympiad geometry \cite{duan2025hageo,zhang2024tonggeometry}. However, these systems still depended on formal or semi-formal problem representations. Newclid addressed this usability issue by proposing a more user-friendly alternative to AlphaGeometry \cite{sicca2024newclid}. These works showed that while geometry theorem proving has advanced significantly, translating natural-language problems into solver-compatible syntax is still a key bottleneck.

\subsection{Challenges in AlphaGeometry Translation}

The task of converting English geometry problems into AlphaGeometry syntax is closely related to auto-formalization. Wu et al. showed that large language models can perform part of this translation process for mathematical statements \cite{wu2022autoformalization}, while Jiang et al. demonstrated that informal proof sketches can help guide formal theorem proving \cite{jiang2022draft}. Benchmarks such as miniF2F have also evaluated IMO Olympiad mathematical reasoning in formal systems such as Lean, Isabelle, and Metamath \cite{zheng2022minif2f}. More recent work on formal mathematical reasoning, such as research related to LeanDojo, ProofBridge, and AlphaProof, further showed the growing importance of connecting natural-language reasoning with formal theorem-proving environments \cite{yang2023leandojo,jana2025proofbridge,hubert2025alphaproof}. However, most of these works focused on general mathematical formalization or proof generation. In contrast, we focus specifically on translating natural-language geometry problems into AlphaGeometry-compatible syntax, where readability, object definitions, givens, and target conclusions must all be correct.

\subsection{LLM Benchmarks and Prompting}

Existing LLM benchmarks evaluated mathematical reasoning or language translation, but not mathematical translation. The MATH dataset and GSM8K benchmark evaluated general mathematical problem solving, while OlympiadBench evaluated Olympiad-level scientific and mathematical reasoning \cite{hendrycks2021math,cobbe2021gsm8k,he2024olympiadbench}. Model-specific work such as DeepSeekMath showed that training on mathematical data can improve competition-math performance, but stronger mathematical reasoning did not necessarily guarantee correct formal translation \cite{shao2024deepseekmath}. For translation prompting, chain-of-thought prompting and self-consistency prompting have shown that LLMs can improve reasoning performance when encouraged to generate intermediate steps or sample multiple reasoning paths \cite{wei2022chain,wang2022selfconsistency}. For AlphaGeometry translation, similar prompting strategies can assist models with improving their translation. We address a specific gap in mathematical reasoning and translation by evaluating LLMs on natural-language-to-AlphaGeometry translation rather than only mathematical problem solving.

\section{NL2AGBench}

\subsection{Benchmark Construction}
\subsubsection{Sourcing Dataset}

NL2AGBench is constructed from geometry problems originally formalized for AlphaGeometry. We begin with the Java Geometry Expert (JGEX) repository, an open-source geometry theorem-proving framework containing hundreds of Olympiad-style geometry problems. Among these problems, the AlphaGeometry project manually formalized 231 instances into its domain-specific language. These examples form the foundation of our benchmark because they enable direct evaluation of automatic translation from English geometry statements into AlphaGeometry-compatible syntax. Unlike general mathematical reasoning datasets, every problem in NL2AGBench contains verified formal specifications that can be executed directly by AlphaGeometry.


\subsubsection{Benchmark Selection}

From the 231 formalized problems, we select 48 representative problems to construct NL2AGBench. Problems were chosen based on two criteria:

\begin{itemize}
    \item Translation diversity: Problems cover a broad range of geometric concepts including circles, cyclic quadrilaterals, angle bisectors, orthocenters, circumcenters, midpoints, reflections, and perpendicular constructions.
    \item Execution efficiency: Problems requiring excessive runtime or computational resources within AlphaGeometry are excluded to ensure scalable evaluation across LLMs.
\end{itemize}

NL2AGBench contains problems (see sample problems 1 and 2 below) spanning varying levels of syntactic complexity and geometric reasoning difficulty while maintaining practical evaluation costs. A visual diagram of 8 sample NL2AGBench problems is shown in Figure 2.


\begin{lstlisting}
Problem 1:
Let ABC be a triangle with circumcenter O. Let H be
the midpoint of BC, and let lines OH and AB 
intersect at D. Through C, draw the line 
perpendicular to CO, and through A, draw the line 
perpendicular to AO; let these two lines intersect 
at E. Prove that points A, O, E, and D lie on a 
circle.

Translation:
a b c = triangle a b c; o = circle o a b c; h = 
midpoint h c b; d = on_line d o h, on_line d a b; e 
= on_tline e c c o, on_tline e a a o 
? cyclic a o e d
\end{lstlisting}

\begin{lstlisting}
Problem 2:
Let ABC be a triangle inscribed in a circle with 
center O. Let M be the midpoint of side BA. Let N be
a point on line OM extended that also lies on the 
circle with center O, so N is the second 
intersection of line OM with the circumscribed 
circle of triangle ABC. Draw line NC. Prove that the 
angle ACN is equal to the angle NCB, that is, prove 
that CN bisects the angle ACB.

Translation:
a b c = triangle a b c; m = midpoint m b a; o = 
circle o a b c; n = on_line n o m, on_circle n o a ?
eqangle c a c n c n c b
\end{lstlisting}



\begin{figure*}[ht]
\centering
\includegraphics[width=0.9\textwidth]{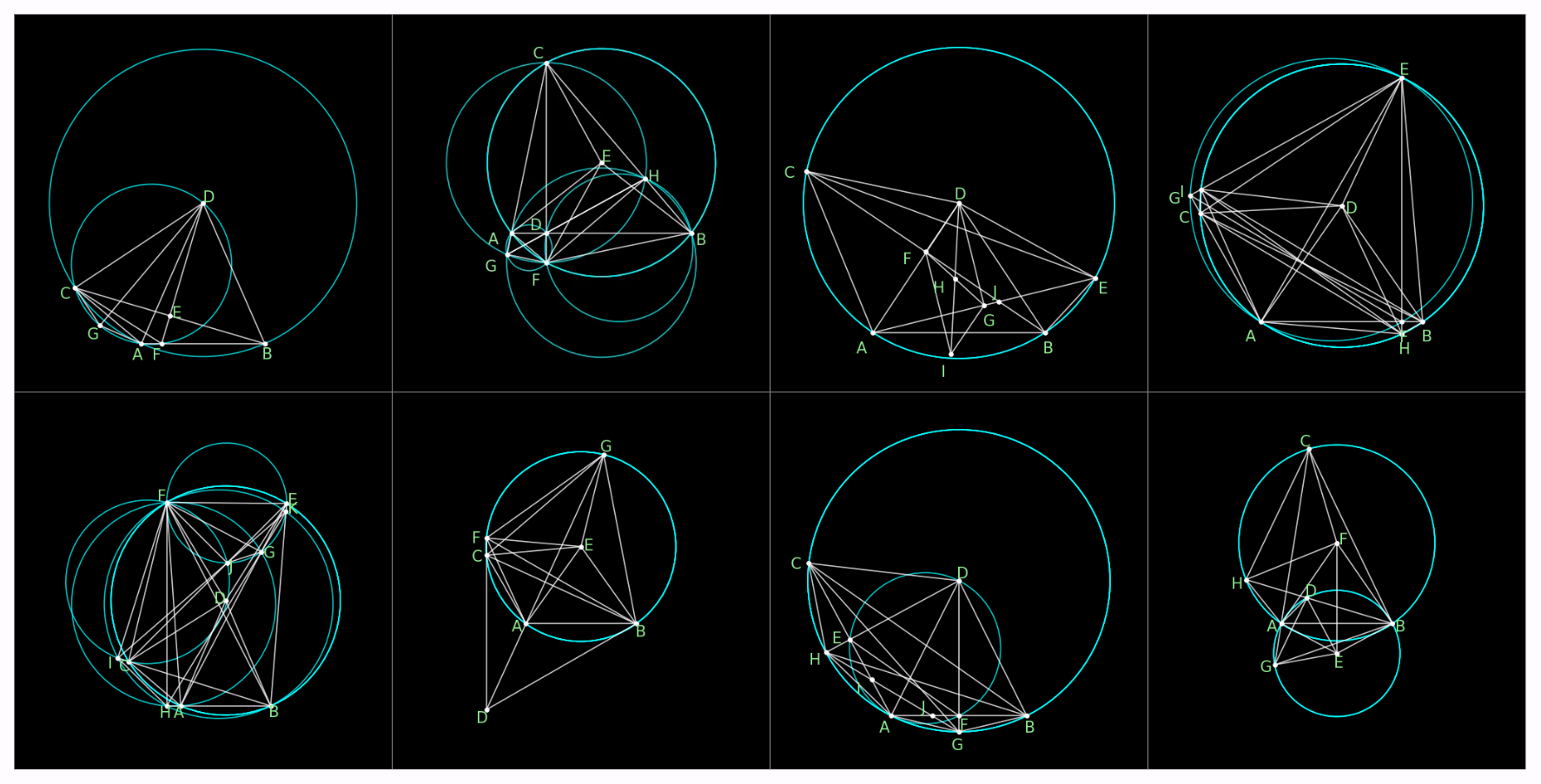}
\caption{Sample NL2AGBench problems. These selected problems encompass a wide range of geometry question types, while still being optimized for runtime and usability efficiency.}
\label{fig:benchmark}
\end{figure*}

\subsubsection{Evaluation Objective}

Given an English geometry problem ($P_{NL}$), the goal is to generate a corresponding AlphaGeometry formalization ($P_{AG}$):

\[
f_\theta(P_{NL}) \rightarrow P_{AG}
\]

where $f_\theta$ denotes an LLM.

The generated translation must satisfy two requirements:

\begin{itemize}
    \item \textbf{Syntactic correctness}: the DSL must conform to AlphaGeometry grammar.
    \item \textbf{Semantic correctness}: the translation must preserve all geometric objects, constraints, and target conclusions contained in the original problem statement.
\end{itemize}

A translation is considered successful if and only if it can be executed by AlphaGeometry without error and produces a valid problem specification.

\subsection{Prompting Strategy}  

To provide models with sufficient knowledge of AlphaGeometry syntax, we employ a prompting framework, which consists of:

\begin{itemize}
    \item official AlphaGeometry syntax definitions
    \item handwritten rules describing clause usage and formatting constraints
    \item the target geometry problem
\end{itemize}

The LLM is instructed to generate a single two-line output containing the problem identifier and its corresponding DSL representation. No explanations or intermediate reasoning are requested. This setup mirrors a realistic auto-formalization scenario in which the model must map natural-language geometry descriptions directly into executable symbolic representations.
\begin{lstlisting}
Using the files above, translate the following 
geometry problem into the AlphaGeometry domain 
language format. Keep the problem title exactly as: 
{problem_id}. Provide ONLY ONE translation. Do not 
include  explanations, commentary, or alternatives 
---just the two-line problem definition (title line 
+ definition line).

\end{lstlisting}
\vspace{-6pt}

\subsection{Translation Verification}
Evaluating translation quality using string similarity alone is insufficient because multiple formalizations may be syntactically different while remaining semantically equivalent. Therefore, NL2AGBench employs an execution-based evaluation protocol. For each generated translation, we execute the resulting DSL specification using AlphaGeometry. The execution outcome is categorized into one of three classes:

\begin{itemize}
    \item \textbf{Successful Translation}: The specification is accepted by AlphaGeometry and executes successfully.
    \item \textbf{Syntax Error}: The specification violates DSL grammar or construction rules.
    \item \textbf{Logic Error}: The specification is syntactically valid but semantically inconsistent with the intended geometry problem.
\end{itemize}

This evaluation framework directly measures whether a generated translation can function as an effective interface between natural-language geometry problems and AlphaGeometry's symbolic reasoning engine.

\section{Error Classification}

A central challenge in evaluating AlphaGeometry formalization is the ambiguity of execution failures. AlphaGeometry frequently reports low-level runtime exceptions that provide limited information regarding the underlying translation mistake. To better understand model behavior, we introduce an execution-based error taxonomy that maps AlphaGeometry failures to interpretable categories.


\subsection{Syntax Errors}
Syntax errors occur when a generated translation violates the structural requirements of the AlphaGeometry domain-specific language. These errors prevent the symbolic engine from parsing or constructing the geometric configuration. Below are the three cases for syntax errors.


\subsubsection{Assertion Error}
\texttt{The AssertionErrors} occur when a construction receives an invalid number of arguments. Such errors typically arise when the model correctly identifies the intended construction but fails to provide all required geometric entities. Examples include incomplete reflection operations, malformed circle constructions, or incorrectly specified auxiliary points. These errors indicate partial understanding of the problem but incomplete knowledge of DSL syntax.


\begin{lstlisting}
Example erroneous translation:
a b = segment; c = midpoint b a; d = reflect c b; e 
= on_circle c a, r_triangle e c d; f = on_tline b a 
b, on_line a e; g = on_line b f, on_line d e ? cong 
e g g f
\end{lstlisting}

In this example, \texttt{AssertionError} is thrown at \lstinline{d = reflect c b} because the reflect construction requires three points: the point that will be reflected and another two points to define the line being reflected upon. Since only two points are provided, the computation results in an error.

\subsubsection{Key Error} A
\texttt{KeyError} occurs when clauses, delimiters, or construction arguments are used improperly, similar to \texttt{AssertionError}. Common causes include:

\begin{itemize}
    \item misuse of special characters (e.g. semicolon, whitespace, proof-operator)
    \item invalid output-variable references
    \item malformed construction declarations
    \item hallucinated DSL clauses
\end{itemize}


\begin{lstlisting}
Example erroneous translation:
b c d a = quadrangle; o = circumcenter b c d a; q 
= midpoint c b; s = midpoint a d; j = midpoint s q; 
m = reflect o j; i = on_line a d, on_line b c ? perp 
s m b c
\end{lstlisting}

AlphaGeometry's DSL allows points to be repeated for readability. For instance, \lstinline{a = midpoint a b c} defines a new point \texttt{a} that is in the middle of segment \texttt{bc}. The point \texttt{a} can be repeated to the right of the clause to show that it is an output point. In the above example, \texttt{KeyError} is thrown at \lstinline{o = circumcenter b c d a}. The circumcenter clause normally takes three points to define the circle in which the circumcenter point is to be constructed. The correct usage including repetition of the output point is \lstinline{o = circumcenter o b c d}. There is a maximum of four points including the repeat to the right of the clause. However, if the output point does not match (\texttt{o} does not match \texttt{b} on the other side in the original example) when the maximum of four points is used, a \texttt{KeyError} is thrown. 

A second source of \texttt{KeyError} is incorrect use of special characters, particularly semicolons and question marks. In the example below, a \texttt{KeyError} occurs before the proof statement \lstinline{foot a d e; ? eqangle} because semicolons may separate clauses but cannot immediately precede a proof clause marked by \lstinline{?}. Similar parsing issues, such as trailing whitespaces, can also trigger \texttt{KeyError}.

\begin{lstlisting}
Example erroneous translation:
a b = segment; c = midpoint b a; d = on_circle c a;
e = on_line a b, on_tline d c d; f = 
foot a d e; ? eqangle f a d a d a b
\end{lstlisting}


The final observed case in which a \texttt{KeyError} occurs is when clauses are misplaced. 

\begin{lstlisting}
Example erroneous translation:
a b c = triangle; d = foot b a c; e = foot a b c; 
f = foot b d e ? 
eqangle a b a d d b d e = eqangle c b c f f b f d
\end{lstlisting}

Here, an extra \lstinline{eqangle} is hallucinated into the proof clause, causing a \texttt{KeyError}.

\subsubsection{Value Error}
\texttt{ValueErrors} occur when geometric objects are defined inconsistently or recursively. Examples include:

\begin{itemize}
    \item circular point definitions
    \item self-referential constructions
    \item invalid proof predicates
    \item incorrect parameter counts
\end{itemize}

These errors indicate that the model generated syntactically plausible structures that violate underlying geometric construction rules. For example, the \texttt{ValueError} is triggered at \lstinline{d = on_circum a b c, on_circum a o d}. Although point \lstinline{d} is being defined, it is also used in the second \lstinline{on_circum} clause. Because \lstinline{d} has not yet been constrained, it cannot be used to determine its own position, resulting in a \texttt{ValueError}.


\begin{lstlisting}
Example erroneous translation:
a b c = triangle; o = circumcenter a b c; d = 
on_circum a b c, on_circum a o d; p = on_circum a 
b c, on_circum a o p; f = foot p a d; g = foot 
p a b; h = foot p b c; e = foot p c d; i = on_line
f g, on_line h e ? cyclic p g i h
\end{lstlisting}


Another case is when the final proof clause has an incorrect amount of parameters. 

\begin{lstlisting}
Example erroneous translation:
a b c = r_triangle; d = foot c a b ? eqangle c a c d 
a d a b b c b d
\end{lstlisting}

The \lstinline{eqangle} clause only takes 8 arguments, the first four defining the first angle and the last four defining the second angle. In this example, \texttt{ValueError} is thrown because there are 12 arguments passed into \lstinline{eqangle}.


\begin{lstlisting}
Example erroneous translation:
a b d = triangle; c = on_tline c a b, on_tline c b a
; d1 = on_line b c, angle_bisector d b c; e = 
midpoint a d1; f = on_line a c, on_line b d1 ? para
c e b d1
\end{lstlisting}

Here, \texttt{ValueError} is thrown at the definition of point c: \lstinline{c = on_tline c a b, on_tline c b a}. The construction \lstinline{on_tline} requires three argument points to establish a point of tangency between a circle and a line. 

\subsection{Logic Errors}
Logic errors occur when the generated translation is syntactically valid but semantically incorrect. Typical causes include:

\begin{itemize}
    \item omitted constraints
    \item incorrect geometric relations
    \item incorrect target predicates
    \item misinterpretation of geometric terminology
\end{itemize}

Logic errors are particularly important because they reveal limitations in mathematical understanding, rather than deficiencies in syntax generation.


\begin{lstlisting}
Example erroneous translation:
a b c = r_triangle a b c ? perp a b b c
\end{lstlisting}

In this example, \lstinline{r_triangle} creates a right triangle with right angle \texttt{CAB}. Therefore, lines \texttt{CA} and \texttt{AB} are perpendicular, not \texttt{AB} and \texttt{BC}. Using the modified ag4masses \cite{tpgh24_ag4masses} system, these logic errors are detected immediately and cause the terminal to freeze. In contrast, the official system continues attempting to solve the problem before eventually freezing as well. Therefore, we terminate any run that exceeds five minutes without producing new output and classify it as a logic error.

\begin{lstlisting}
I0529 12:26:07.656355 140703903308672 graph.py:498]
test_problem
I0529 12:26:07.656621 140703903308672 graph.py:499]
a b c = r_triangle a b c ? perp a b b c
*terminal freezes beyond this point
\end{lstlisting}

\subsection{Significance of the Taxonomy}

Our taxonomy provides a standardized framework for analyzing failures in AlphaGeometry auto-formalization. Rather than reporting only overall success rates, the taxonomy enables fine-grained diagnosis of whether model failures arise from syntax generation, geometric interpretation, or semantic reasoning. This distinction is critical for not only understanding the errors designing future mitigation strategies and improving LLM-based mathematical formalization systems.

\subsection{Baseline Model Performance Results}

\begin{table}[h]
\centering
\caption{Model Performance (Zero-Shot)}
\label{tab:results}
\begin{tabular}{lc|lc}
\hline
\textbf{Model} & \textbf{Correct} & \textbf{Model} & \textbf{Correct} \\
\hline
\multicolumn{2}{l|}{\textit{Closed Source}} & \multicolumn{2}{l}{\textit{Open Source}} \\
\hline
GPT-5.4      & 75.00\% & qwen3:235B        & 12.50\% \\
Gemini 3.1   & 62.50\% & Llama3.1:70B      & 0\%     \\
Grok-3       & 52.08\% & Llama3.2:3B       & 0\%     \\
Sonnet 4.6   & 37.50\% & Mistral:7B        & 0\%     \\
gpt-4o:8B    & 4.17\%  & nemotron-3-nano:4B & 0\%    \\
\hline
\end{tabular}
\end{table}

\section{Improving Translation Performance}

Our baseline evaluation reveals a substantial performance gap between frontier closed-source models and open-source alternatives. The hatched data in Figure 3 illustrates the performance of these models in zero-shot circumstances and serve as a reference point. While large proprietary models frequently generate executable AlphaGeometry translations, open-source models often fail due to syntax violations, omitted constraints, or incorrect geometric interpretations. To better understand how translation performance can be improved, we investigate three mitigation strategies: few-shot prompting, human-guided correction and supervised fine-tuning.




\subsection{Few-Shot Prompting}

\begin{figure*}[h]
    \centering
    \includegraphics[width=.9\linewidth]{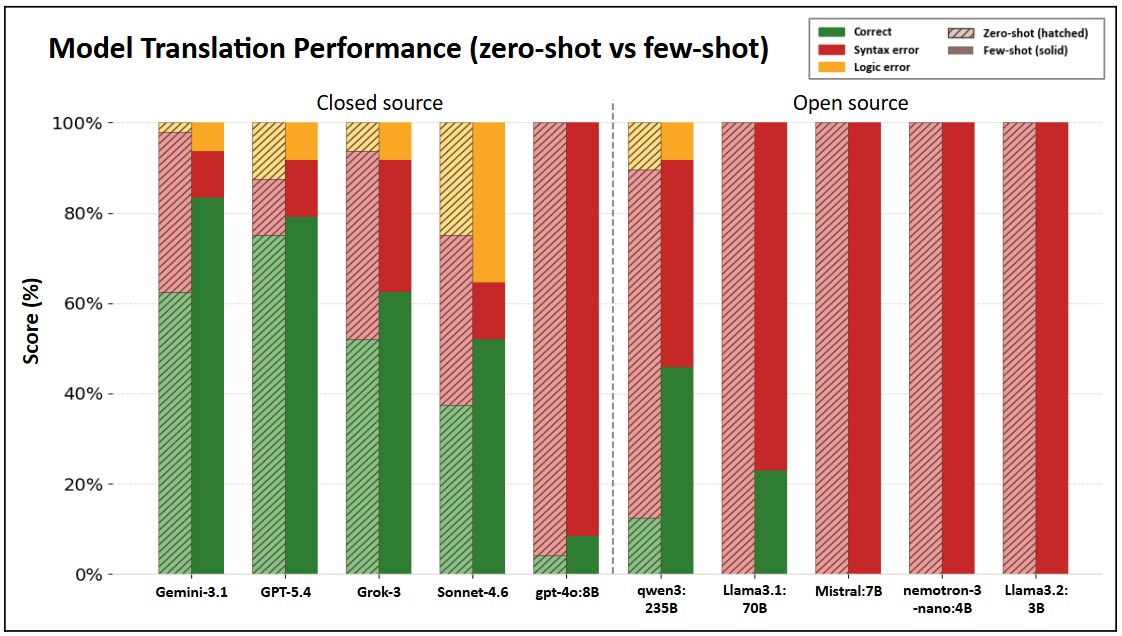}
    \caption{Model performance for zero-shot and few-shot prompting. }
    \label{fig:results}
\end{figure*}

Few-shot prompting serves as the primary intervention strategy. Each model receives 54 reference examples consisting of paired natural-language geometry problems and AlphaGeometry translations. These examples expose the model to common geometric constructions, DSL syntax patterns, and proof predicates. Compared with zero-shot prompting, few-shot prompting consistently improves executable translation rates across all evaluated models. 

Below are two samples of the multi-shot example translations that are supplied to LLMs. The first section of the example depicts the English version of the problem, while the second is the manually-verified DSL translation. 

\begin{lstlisting}
# Example 1
Let ABC be a triangle with AB < AC and circumcenter 
O. The angle bisector of angle BAC meets BC at D.
The line through D perpendicular to BC meets AO at 
X. Y is the midpoint of AD. Prove that B, C, X, Y 
are concyclic.
    
a b c = triangle; o = circumcenter a b c; d =
on_line b c, angle_bisector b a c; x = on_line a o,
on_tline d b c; y = midpoint a d ? cyclic b c x y

# Example 2
Let ABC be a triangle with AB=AC and M the midpoint 
of BC. P is a point with PB<PC and PA parallel to BC.
X,Y are on lines PB,PC with B on PX, C on PY, and 
angle PXM = angle PYM. Prove APXY is cyclic.

a b c = iso_triangle; m = midpoint b c; p = on_pline 
a b c; x = on_line p b; y = on_line p c, eqangle3 y
p m x m p ? cyclic a p x y
\end{lstlisting}

Figure 3 shows that few-shot prompting improves translation accuracy across nearly all models. Among closed-source models, Gemini-3.1 achieves the best performance, increasing from 62.5\% accuracy in the zero-shot setting to 83.3\% with few-shot prompting, a gain of about 21 percentage points. Grok-3 improves from 52.1\% to 62.5\% (+10 points), while Sonnet-4.6 rises from 37.5\% to 52.1\% (+14 points). GPT-5.4 starts at a relatively high 75.0\% accuracy and reaches 79.2\% with few-shot examples (+4 points). In contrast, GPT-4o:8B remains slightly improved, improving only from 4.2\% to 8.3\%.

The largest relative gains are observed among open-source models. Qwen3:235B improves from 12.5\% to 45.8\% accuracy, a gain of around 33 percentage points and more than a threefold increase. Llama3.1:70B rises from 0\% to 23\%, while Mistral:7B, Nemotron-3 Nano:4B, and Llama3.2:3B fail to produce any correct translations in either setting. The improvement is particularly pronounced for syntax-related errors, suggesting that a substantial fraction of translation failures arise from unfamiliarity with AlphaGeometry's DSL rather than a complete lack of geometric understanding.

Overall, few-shot prompting is the most effective and scalable approach, consistently reducing error rates and improving accuracy across models without requiring retraining. However, a substantial performance gap remained between the strongest closed-source models and most open-source alternatives.

\subsection{Human-Guided Hints}

Our next strategy investigates whether targeted human guidance can help LLMs recover from translation failures. Unlike few-shot prompting and fine-tuning, which provide general knowledge about AlphaGeometry syntax, this approach introduces problem-specific information after an initial translation has been generated.

As illustrated in Figure \ref{fig:diagram_hint}, we employ a two-stage human-guided correction framework.

\subsubsection{Stage 1: Diagram-Based Hints}

In the first stage, the model is provided with a geometric diagram corresponding to the target problem and is asked to re-evaluate its translation. The diagram supplies spatial information that may be difficult to infer reliably from text alone, including collinearity relationships, circle memberships, and auxiliary constructions.

Many Olympiad geometry problems implicitly assume information that is visually obvious from the accompanying figure but only partially specified in the textual description. By incorporating the diagram shown in Figure~4, the model gains access to additional geometric context that can reduce ambiguity during formalization. This stage is effective when the original translation omits geometric constraints or incorrectly interprets relationships between points, lines, and circles.

\subsubsection{Stage 2: Error-Specific Hints}

After examining the revised translation, a second round of guidance is provided when necessary. Rather than supplying additional geometric context, these hints directly target observed translation errors. As shown in the lower portion of Figure~3, the hint may identify a specific issue such as an incorrect number of clause parameters, misuse of a geometric predicate, or an invalid construction. The model is then asked to revise its translation while preserving the remaining structure. This stage functions as a targeted correction mechanism. Instead of requiring the model to rediscover the source of failure, the hint narrows the search space and focuses attention on the problematic portion of the formalization.

\begin{figure}
    \centering
    \includegraphics[width=.9\linewidth]{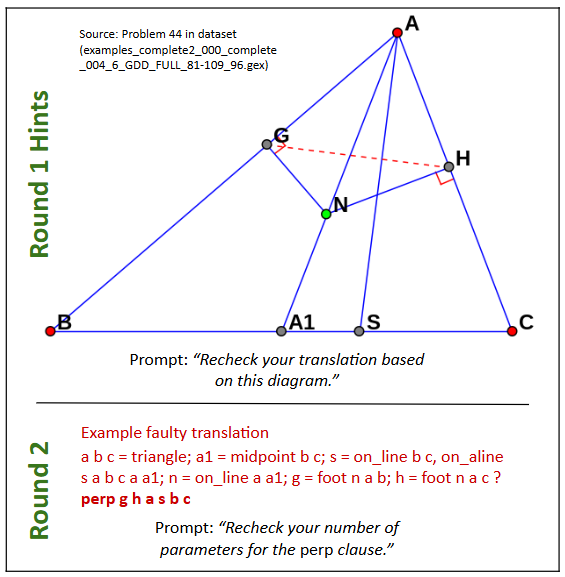}
    \caption{Two-stage human-guided correction framework. In Stage 1, the model is provided with a geometry diagram to supply additional structural context and reduce ambiguity in the natural-language problem statement. In Stage 2, targeted error-specific hints are provided to correct localized translation mistakes identified in the generated AlphaGeometry formalization.}
    \label{fig:diagram_hint}
\end{figure}
\subsubsection{Effectiveness and Limitations}

Human-guided hints improve translation quality for strong frontier models, particularly when the initial translation is close to correct. In many cases, the models demonstrate sufficient geometric understanding to generate an accurate formalization once a specific ambiguity or error has been identified. Figure \ref{fig:diagram_hint} highlights the complementary roles of the two stages. The first stage addresses missing geometric information through diagram-based reasoning, while the second stage addresses localized translation mistakes through targeted corrective feedback. Together, they form an refinement process that improves executable translation rates without modifying model parameters. Specifcially, the hint-based methodology improves translation performance across all evaluated models: GPT-5.4 by 14.29\%, Gemini 3.1 by 14.63\%, Claude Sonnet 4.6 by 26.47\%, and Grok-3 by 32.14\%. These gains show that targeted human guidance can substantially improve translation rates, especially for cases involving recoverable formalization errors.

Despite their effectiveness, human-guided hints have limited scalability. Diagram interpretation requires additional problem resources, and error-specific hints require manual inspection of model outputs. Consequently, this approach is best viewed as a diagnostic and correction framework rather than a fully automated solution for large-scale benchmark evaluation.

\subsection{Supervised Fine-Tuning}

We additionally investigate whether fine-tuning can improve translation performance for smaller open-source models. For this experiment, we utilize HAGeo-409, a human-verified collection of geometry problems paired with AlphaGeometry formalizations. Models are fine-tuned to generate DSL representations directly from natural-language inputs. Fine-tuning improves adherence to AlphaGeometry syntax and reduces several categories of grammar-related errors. However, improvements in executable translation accuracy remain limited.
Qualitative inspection reveals that many fine-tuned models learn surface-level DSL patterns without reliably capturing the geometric semantics required for correct formalization. In several cases, models generate syntactically plausible but semantically incorrect constructions, indicating that formalization requires deeper geometric reasoning than simple syntax imitation. These findings suggest that the primary bottleneck for smaller models is not solely DSL familiarity but also insufficient reasoning capacity.

\section{Conclusion and Future Work}

This paper investigates a key bottleneck in automated geometry theorem proving: translating natural-language geometry problems into the formal language required by AlphaGeometry. To address this challenge, we introduce NL2AGBench, a benchmark specifically designed to evaluate the ability of large language models to translate English geometry problems into AlphaGeometry-compatible formal representations. 



We systematically assess both open-source and closed-source LLMs across a diverse collection of geometry problems. Our results reveal a substantial gap between frontier proprietary models and current open-source alternatives, suggesting that successful formalization requires not only syntactic mastery of the target language but also accurate geometric interpretation and constraint preservation. Through detailed error analysis, we construct a taxonomy of AlphaGeometry translation failures, categorizing them into syntax errors and logic errors. This taxonomy provides a practical framework for understanding model weaknesses and diagnosing failure modes in future auto-formalization systems.

We further explore multiple mitigation strategies, including few-shot prompting, supervised fine-tuning, and human-guided hints. Among these approaches, few-shot prompting is the most scalable and consistently effective strategy, while human-guided correction demonstrates that some translation failures arise from localized ambiguities rather than a complete lack of geometric reasoning capability. These findings suggest that improvements in mathematical formalization may require tighter integration between language understanding, geometric reasoning, and symbolic verification.

Future work could extend NL2AGBench to additional theorem-proving systems, incorporate geometric diagrams through multimodal formalization, develop specialized auto-formalization models, and integrate symbolic feedback for iterative self-correction.

\bibliographystyle{IEEEtran}
\bibliography{references}

\vspace{12pt}


\end{document}